\documentclass[letterpaper, 10 pt, conference]{ieeeconf}  

\IEEEoverridecommandlockouts                              

\makeatletter
\let\NAT@parse\undefined
\makeatother
\usepackage[numbers]{natbib}

\usepackage[pdftex]{graphicx}
\usepackage{amsmath}
\usepackage{amssymb}  
\usepackage{subfig}
\usepackage{multirow}
\usepackage{array,booktabs}
\usepackage{diagbox}
\usepackage{balance}
\usepackage[ruled]{algorithm}
\usepackage{algpseudocode}
 \usepackage[final]{hyperref}
 \hypersetup{
     colorlinks=true,
     linkcolor=blue,
     filecolor=magenta,
     urlcolor=blue,
     citecolor=black
 }
\usepackage[font=small]{caption}
\usepackage{./rpm_packages/rpm_SIunits}
\usepackage{./rpm_packages/rpm_acronyms}
\usepackage{./rpm_packages/rpm_math}
\usepackage{./rpm_packages/rpm_misc}

\usepackage{soul,color}
\usepackage{lipsum}

\usepackage{xcolor}

\newcommand{\bl}[1]{{\textcolor{black}{#1}}}

\DeclareMathOperator*{\argmin}{argmin}

\usepackage{tikz} 

\title{\LARGE \bf
RaPlace: Place Recognition for Imaging Radar\\ using Radon Transform and Mutable Threshold
}     

\author{Hyesu Jang${}^{1}$, Minwoo Jung${}^{1}$,  and Ayoung Kim${}^{1*}$
\thanks{$^\dagger$\bl{This work was jointly supported by the NRF (RS-2023-20241758) and the MOTIE (20024355), Korea.}}
\thanks{$^{1}$H. Jang, M. Jung, and A. Kim are with the Department of Mechanical Engineering, SNU, Seoul, S. Korea {\tt\small [dortz, moonshot, ayoungk]@snu.ac.kr}}%
}

\begin{document}

\maketitle
\thispagestyle{empty}
\pagestyle{empty}

\begin{abstract}

Due to the robustness in sensing, radar has been highlighted, overcoming harsh weather conditions such as fog and heavy snow. In this paper, we present a novel radar-only place recognition that measures the similarity score by utilizing Radon-transformed sinogram images and cross-correlation in frequency domain. Doing so achieves rigid transform invariance during place recognition, while ignoring the effects of radar multipath and ring noises. In addition, we compute the radar similarity distance using mutable threshold to mitigate variability of the similarity score, and reduce the time complexity of processing a copious radar data with hierarchical retrieval. We demonstrate the matching performance for both intra-session loop-closure detection and global place recognition using a publicly available imaging radar datasets. We verify reliable performance compared to existing stable radar place recognition method. Furthermore, codes for the proposed imaging radar place recognition is released for community \bl{https://github.com/hyesu-jang/RaPlace}.

\end{abstract}
\section{Introduction}
\label{sec:intro}

Radar is an essential range sensor for mobile robotics, playing an important role in safety and environment detection. Although camera and \ac{LiDAR}-based scene understanding \bl{are widely adopted in autonomous vehicles}, these sensors may be limited in harsh weather, such as thick fog, sandstorms, or heavy snow. On the contrary, radar inherently surmounts the abovementioned challenges with a more extended wavelength signal. Since radar has a long-range capability and penetrates small particles, sensor data is impervious to occlusion. Despite these advantages, radar place recognition is at an incipient stage compared to camera and LiDAR-based place recognition due to the major bottleneck in radar-based navigation originating from peculiar radar characteristics. 

Radar image pixel values are represented with \ac{RCS}, radar wave reflection criterion. \ac{RCS} differs from every medium that radar wave reflects, which can be used as feature points. Unfortunately, intensive noise level and false vivid objects also possess high \ac{RCS} values, requiring the ambiguous points removal. Hence, the main challenge of imaging radar has been differentiating noise and feature points, obtaining valid features for robot navigation.


\begin{figure}[!t]
    \centering
    \includegraphics[width=\columnwidth]{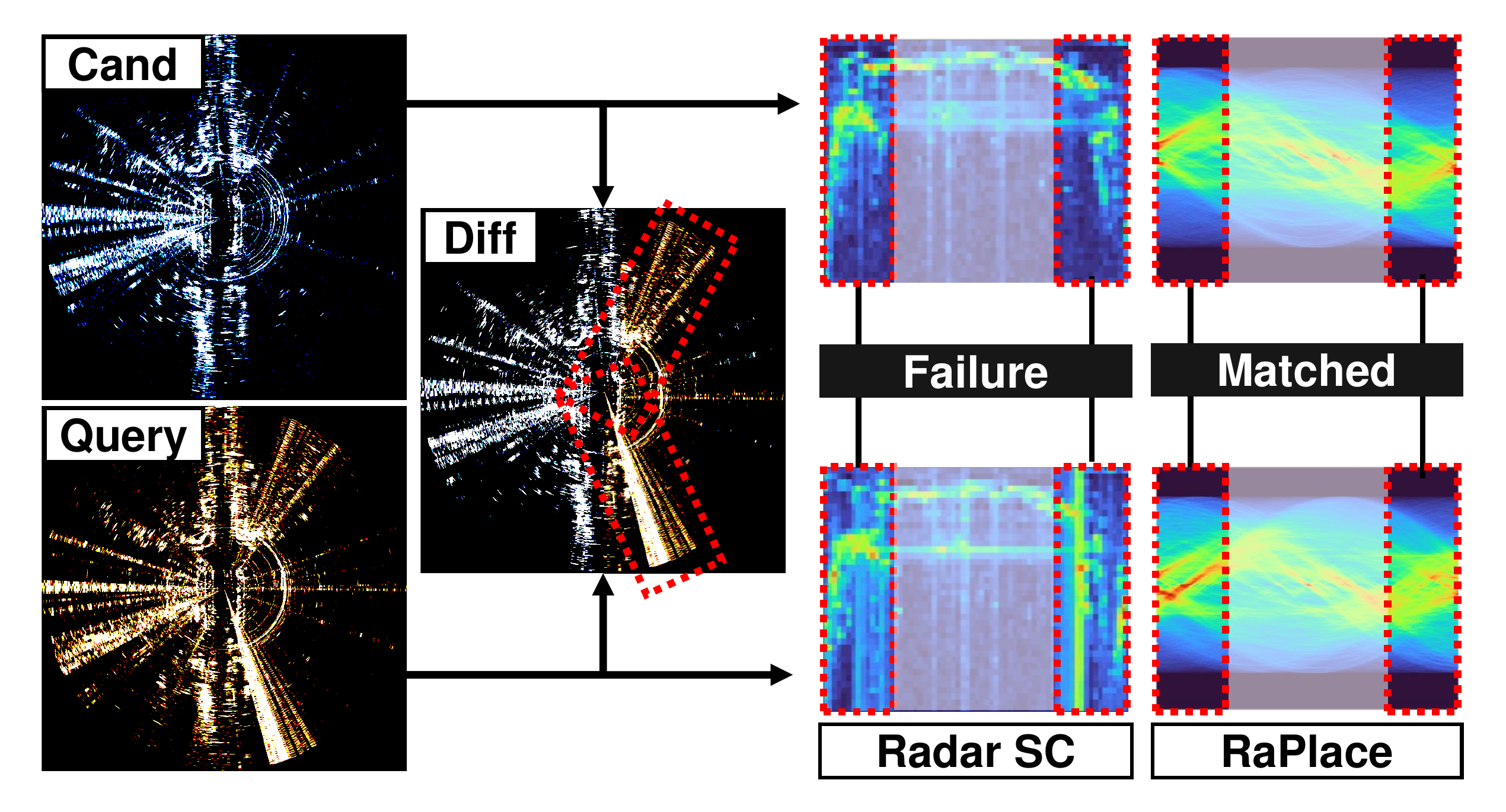}
    \caption{A failure case in radar place recognition using the \textit{Radar Scan Context} approach is observed due to the presence of vivid radar multipath noise and ring noises that appear as strong \ac{RCS} signals. This hinders the ability of the \textit{Radar Scan Context} approach to accurately distinguish between the noise and the feature points. However, \textit{RaPlace} successfully overcomes the challenges posed by these noises, enabling effective place recognition performance.}
    \label{fig:intro}
    \vspace{-5mm}
\end{figure}

The key to imaging radar place recognition is utilizing the image information and \ac{RCS}. Therefore, extracting meaningful information from the polar radar images and simultaneously searching for the candidates with reliable accuracy is compulsory. Some tackled this issue by adopting LiDAR-based methods. For example, \citet{hong2020radarslam} and \citet{kim2020mulran} replaced LiDAR height information with radar \ac{RCS} to implement radar place recognition methods.
The other approaches were learning-based methods~\cite{de2020kradar++,gadd2020look}, demonstrating reliable results by improving the well-known visual place recognition, NetVLAD~\cite{arandjelovic2016netvlad}. 

While these methods are groundbreaking in imaging radar place recognition, they do not contain a metric algorithm that takes into account radar noise and \ac{RCS} properties.
Despite being ranging sensors, LiDAR-based methods are insufficient in comprehensively encompassing radar data and its accompanying noise.
The learning-based methods also exhibit these limitations, and treat radar images as camera images or LiDAR point clouds.
Additionally, learning-based imaging radar place recognition approaches are not publicly accessible, thereby algorithm reproducing for verification is challenging.
While prior research has established the reliability, it is obvious that a readily available metric algorithm is required.

For the recent work, \citet{lu2022one} proposed the \ac{RT}-based \ac{LiDAR} place recognition that is both rotation and translation invariant. This work motivated us in that the RT was originally designed to analyze tomography images rather than scanned pointcloud images.
In contrast to LiDAR, radar data shares certain similarities with tomography images, being generated by rays that penetrate the surrounding environment.
Furthermore, tomography images obtained from X-ray imaging may exhibit undesired effects such as radiation saturation and detector noise, which are akin to the noise observed in radar images. Hence, we adopt the \ac{RT} to analyze the radar image. Additionally, since \ac{RT}-applied data forms the basis for noise post-processing, applying \ac{RT} to radar images provides a similar advantage as shown in \figref{fig:intro}.



In this paper, we address the imaging radar-based place recognition method using \ac{RT}, with algorithmically congenial to radar image processing. Instead of eliminating the ambiguous points, we propose to stifle the influence of the futile points. The proposed contributions of our paper are listed below.

\begin{itemize}
    \item \textbf{A new paradigm for the radar place recognition}
    The proposed method is a rotational and translational invariant high-precision place recognition method for radar. The addressed approach can fill up the lacuna in reverse direction driving. To adapt the LiDAR-based method to radar, we remodeled existing similarity distance scoring algorithm and adopted hierarchical candidate extraction with data abridgment.

    \item \textbf{Non-learning radar place recognition criteria for wide tailorability.} 
    The existing criteria for radar place recognition are scarce, a confident localization tool for radar \ac{SLAM} is required. The proposed approach affords high-accuracy place recognition results with low computational complexities \bl{without needing a GPU}.

    \item \textbf{Open-source radar place recognition code}
    Currently few radar place recognition methods are publicly available. We release the code for radar place recognition.
\end{itemize}

\section{related work}
\label{sec:relatedwork}


\subsection{Place Recognition using Cameras and LiDARs}



\subsubsection{Visual Place Recognition using RGB Cameras}

Visual place recognition studies frequently utilize feature-based descriptors~\cite{ng2003sift,rosten2008faster,rublee2011orb}. Some recent studies have leveraged specialized feature-based searching techniques. For example, \citet{arandjelovic2016netvlad}  introduced a learning model architecture for visual place retrieval that combined a CNN architecture with a generalized vector of locally aggregated descriptors.
\citet{qazanfari2019content} employed the HSV color model and image histogram as the basis for image retrieval, utilizing color discrepancies and image histogram as the retrieval features. \citet{panek2022meshloc} generated a 3D mesh by leveraging 2D image depth and employed it for place recognition. While these techniques have yielded promising results, they remain constrained by issues stemming from the camera's inherent limitations, such as variations in illumination and challenging weather conditions. 
Notably, rendering techniques are unsuitable for 2D imaging radar.

\subsubsection{LiDAR Place Recognition}

Although numerous place recognition studies have employed visual sensors, utilizing \ac{LiDAR} is also prevalent. \citet{kim2018scan} proposed the \textit{Scan Context} algorithm, a versatile place recognition approach for LiDAR data. The cosine similarity between two context representations of the data is used to compute the similarity distance between two images. An improved version of this algorithm, \textit{Scan Context++}~\cite{kim2021scan}, was introduced to achieve translational invariance by incorporating Cartesian context and to enhance performance through the use of hierarchical clustering.

\citet{uy2018pointnetvlad} improved the network architecture of PointNet~\cite{qi2017pointnet} and NetVLAD~\cite{arandjelovic2016netvlad} to conduct place recognition in the pointcloud data. Authors utilized descriptors with lazy triplet and quadruplet loss that were found to effectively distinguish pointcloud data, resulting in a high retrieval rate.


\citet{ding2022translation} and \citet{lu2022one} employed \ac{RT} and \ac{DFT} to compress image and implement a global searching algorithm. Especially, \cite{lu2022one} derived the rotation and translation invariance of the proposed place retrieval method and estimated the orientation and translation differences. The extension version of \ac{RT}-based LiDAR place recognition \cite{lu2022one} was recently addressed by \citet{xu2022ring++}, which features the application of multiple sensors. 

We conduct an adaptation process of \ac{RT}-based algorithm for radar images. As demonstrated above, \ac{RT} yields useful elements for reducing noise in tomography images with saturation. We address the issue of radar-induced noise by employing \ac{RT}, where radar images are utilized as tomography images.
To facilitate this adaptation, we conduct preprocessing on the radar images \bl{to address their inherent limitations and to produce} a highly accurate query-pursuant scoring descriptor.

\subsection{Place Recognition with Radars}
Radar sensors can be divided into two types: (\textit{i}) single-chip automotive radar, which provides point clouds with velocity value, and (\textit{ii}) scanning imaging radar, which generates high-resolution image data. 

Automotive radar, also known as single-chip radar, is a commonly used sensor for vehicles. As a result of the limited density of radar point cloud data, the research in the area of localization \cite{ward2016vehicle, schuster2016robust, park2019radar, lu2020milliego} has advanced further than that in place recognition. Recently, \citet{cait2022autoplace} demonstrated the \ac{SOTA} outcomes in automotive radar place recognition. The authors conducted radial velocity-based dynamic points removal to produce stable candidates, encoded spatial and temporal information for retrieval, and enhanced accuracy through RCS histogram reranking.



\begin{figure*}[!t]
    \centering
    \includegraphics[width=0.9\textwidth]{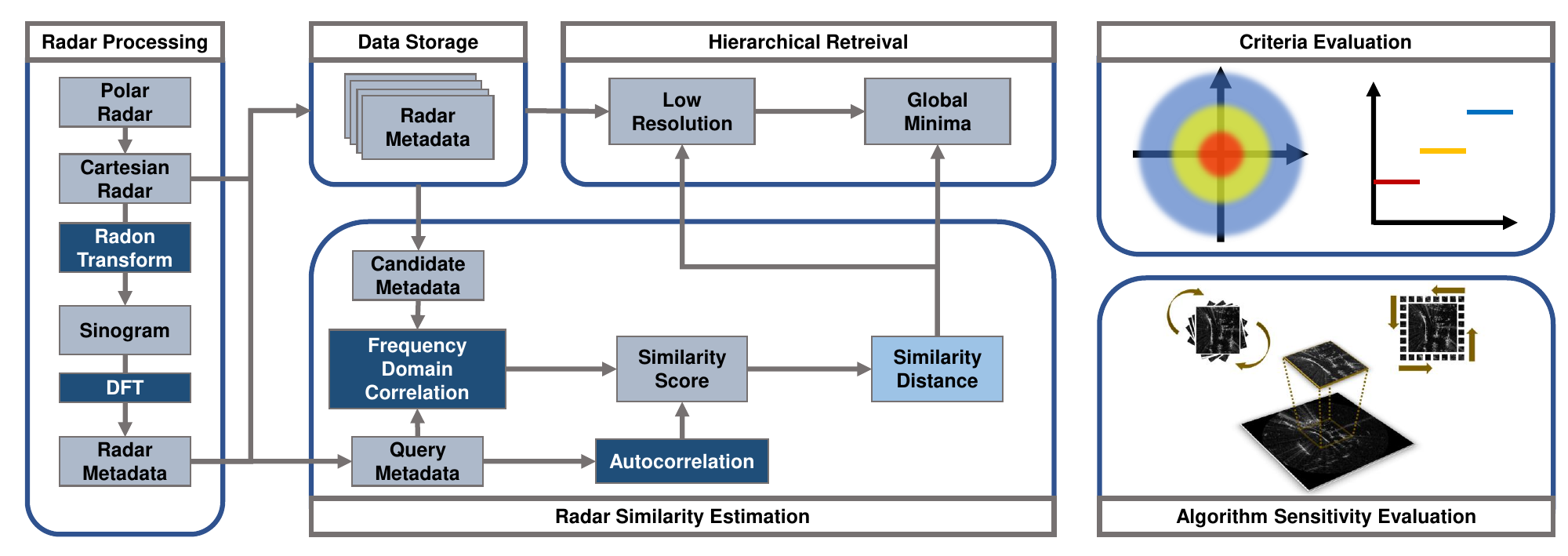}
    \caption{The framework for RaPlace and overview of ablation studies. Initially, processed radar data is transformed into a query data and is stored in a metadata format. Candidates from metadata are evaluated with frequency domain correlation estimation. The resulting correlation scores represent the degree of similarity between the candidates and the query data, and the auto-correlation score provides a mutable threshold for determining the similarity distance. The complete framework is executed on low-resolution images and subsequently extended to high-resolution images to explore the global minima using the abovementioned steps. Ablation studies are carried out to assess the performance of the system, with a focus on criteria and sensitivity evaluation.}
    \label{fig:pipeline}
    \vspace{-5mm}
\end{figure*}

In contrast to the sparse implementation of automotive radar, scanning radar is more frequently employed for the kidnapped vehicle problem.
\citet{kim2020mulran} presented a ranging sensor dataset with radar place recognition evaluation. Instead of using LiDAR elevation values, they utilized radar RCS values to create a radar scan context. \citet{tang2020rsl} attempted to localize satellite images using radar image data. They estimated the rotation value between the radar and satellite images, and constructed a neural network to estimate the translation value and calibrate the images. \citet{suaftescu2020kidnapped} proposed a learning-based approach for topological localization to achieve metric pose estimation. The authors leveraged the fact that the position information of the polar radar image is independent of the heading. \citet{wang2021radarloc} addressed a 6-DOF global vehicle pose estimation network, which employed a self-attention module for localization based on radar images. \citet{yin2021rall} demonstrated multi-modal sensor localization using radar on LiDAR. They used neural network-based data embedding and a differentiable Kalman filter to achieve multi-modal localization.

As summarized above, research on radar place recognition typically relies on neural networks that are challenging to reproduce. Furthermore, the solutions to the kidnapped problem using simple metric algorithms are insufficient. We propose an imaging radar-based non-learning approach to place recognition that is both resilient to noise and capable of \bl{overcoming the inherent limitations} of radar sensors.
\section{Method}
\label{sec:method}

Being a non-learning-based method, we seek a descriptor that can differentiate between candidates analogous to the query data. Our proposed model aims to identify places that include the vicinity of the target location, thus extending the scope of detection and smoothing the disclosure of identical places.
\figref{fig:pipeline} depicts the entire pipeline for our place recognition algorithm. We generate backward-warping Cartesian images from the polar image data and obtain sinogram images using the \ac{RT} on the Cartesian radar data. These acquired sinograms are then converted into FFT images, and utilized as candidates. To determine the similarity score, we adopt the cross-correlation in frequency domain and mutable threshold for all previous keyframes when query images are published. The candidate location is determined as the keyframe with the lowest distance, assessed from the similarity scores. The following subsections provide a detailed account of the procedures.

\subsection{Radar Data Preparation}
\begin{figure}[!t]
    \centering
    \includegraphics[width=\columnwidth]{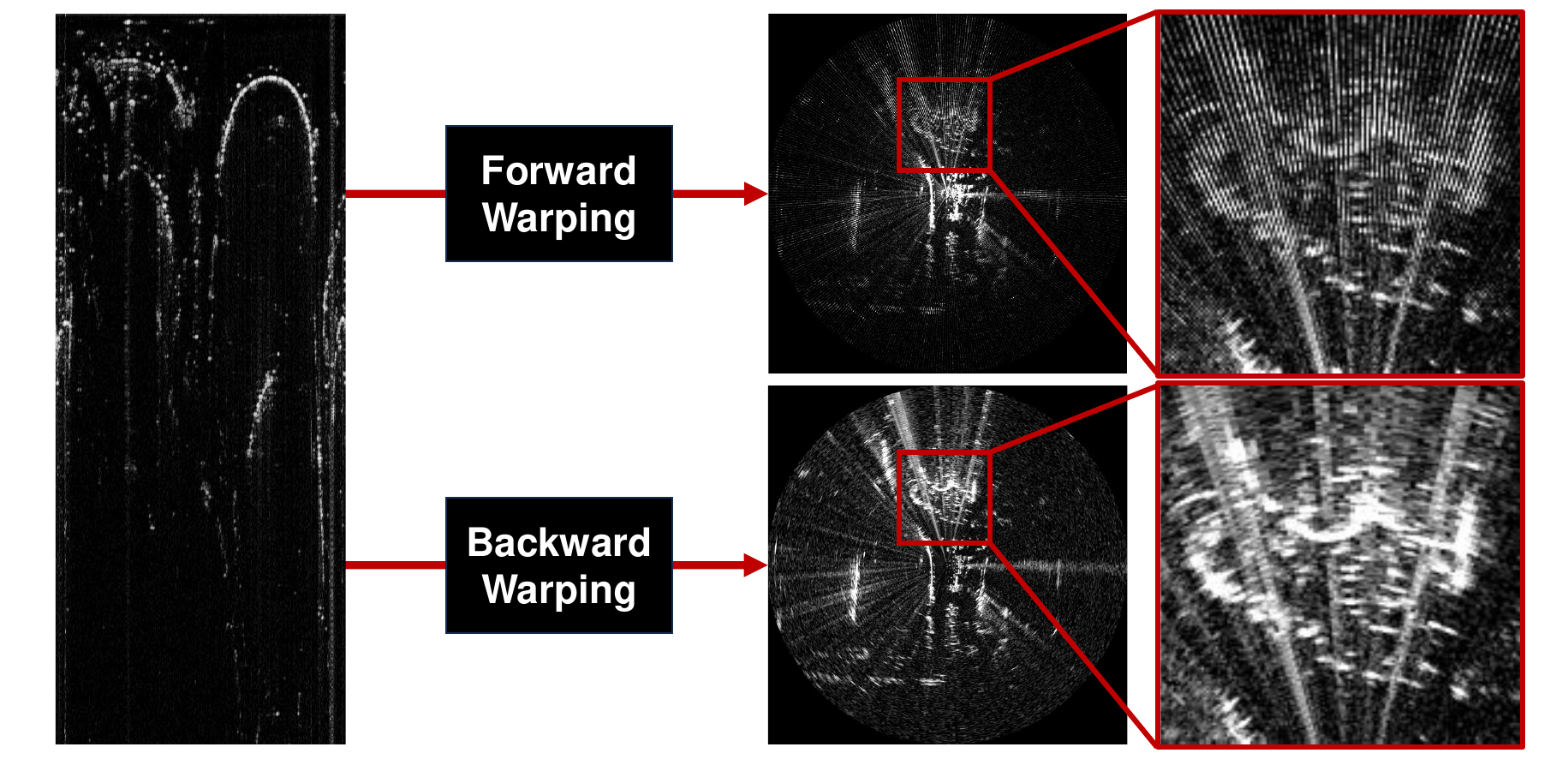}
    \vspace{-1mm}
    \caption{\bl{The outcome of Cartesian imagery varies according to the warp methodologies. The process of forward warping leads to the generation of empty spaces amongst pixels. Contrastingly, backward warping exhibits all-occupied pixel spaces.}}
    \label{fig:warping}
    \vspace{-5mm}
\end{figure}

FMCW radar generates 4Hz polar radar images with RCS values. Polar images are irrelevant to rotational changes however, tracing translational variations is challenging. \bl{To account for translational differences, Cartesian images are required. Unlike LiDAR bird-eye-view images, radar can penetrate obstacles, which allows \ac{RCS} information to be fully captured up to the range limit. To fully utilize this characteristics, we convert the polar images using a backward warping technique. As illustrated in the \figref{fig:warping}, the process of forward warping directly registers the polar pixel value in Cartesian space. As a consequence of this approach, images manifest as a pattern of discrete rays, hindering the complete utilization of the converted image due to the presence of vacant pixels. On the contrary, backward warping initiates with an empty Cartesian space and progressively populates all the xy pixels corresponding to the closest polar pixel. This conversion offers substantial assistance in pixel-level analysis of translational disparities.}

\subsection{Radon Transform for Radar}
\begin{figure}[!t]
    \centering
    \vspace{-3mm}
    \includegraphics[width=\columnwidth]{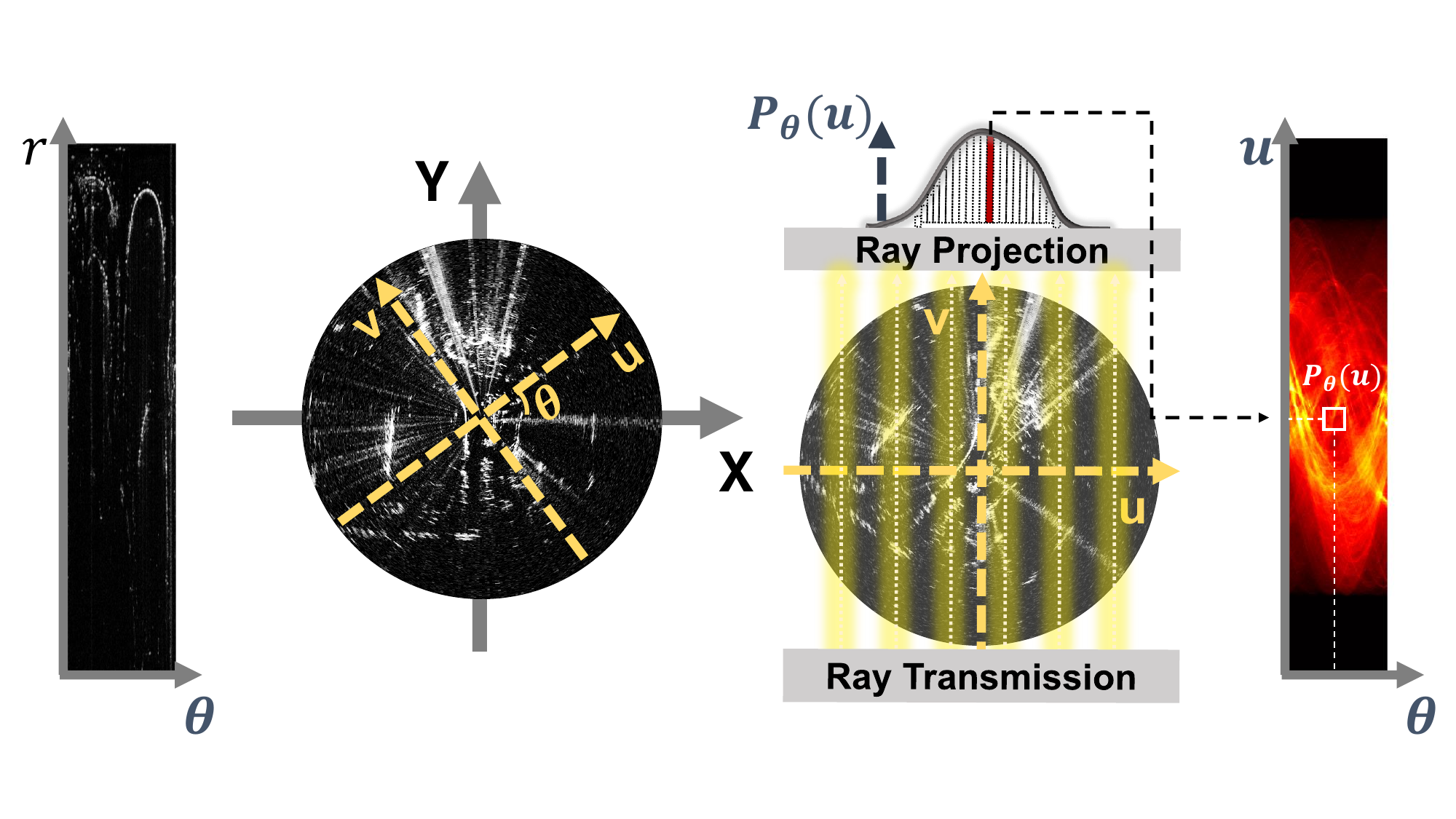}
    \vspace{-8mm}
    \caption{The fundamental concept behind the \ac{RT}. For a given input of a raw radar image, there exists a rotating coordinate system (u,v) that propagates a ray along the v axis. These rays penetrate the radar contents and generate projection vectors that ultimately become the column vectors of the sinogram. }
    \label{fig:radon}
    \vspace{-5mm}
\end{figure}

Summarizing the image data is a crucial step in generating a descriptor. The Radon transform is a commonly used image processing and restoration method in medical image treatment. The representative utilization is \ac{CT}, where rotating light emissions generate the projection vectors for every angle. The column-wise enumeration of these vectors forms a sinogram that encapsulates the entire image data. We decide to employ this image data compression technique for our robotics problem. \figref{fig:radon} is a schematized image of \ac{RT} and sinogram. The relation between $(x,y)$ and $(u,v)$ for the varying coordinate can be expressed using the rotational transform.

\begin{equation}
\begin{bmatrix}
x\\y
\end{bmatrix}
 = 
\begin{bmatrix}
\cos{\theta} & -\sin{\theta}\\
\sin{\theta} & \cos{\theta}
\end{bmatrix}
\begin{bmatrix}
u\\v
\end{bmatrix}
\end{equation}
\bl{We use line integration to measure the image density $P_{\theta}(u)$. Densities are calculated with the rays penetrating the radar detections.}
\begin{equation}
\centering
\begin{split}
P_{\theta}(u) &= \int_{}^{} f(x,y) \,dv \\
& = \int_{}^{} f(u \cos{\theta}-v\sin{\theta},u\sin{\theta}+v\cos{\theta}) \,dv
\end{split}
\end{equation}
The two-dimensional slice that represents the data penetration is denoted by $f(x,y)$. Expanding the equation from a line to the entire area allows us to generate a sinogram image $\mathcal{S}(\theta,u)$, the result of the \ac{RT}. The total densities along the v-axis direction compose the compressed data.
\begin{equation}
\mathcal{S}(\theta,u) =
\begin{bmatrix}
P_{0}(l_{min}) & \cdots & P_{\pi}(l_{min})\\
\vdots & \ddots & \vdots\\
P_{0}(l_{max}) & \cdots &P_{\pi}(l_{max})\\
\end{bmatrix}
\vspace{2mm}
\end{equation}

The non-zero boundary of $P_{\theta}(u)$ is denoted by $l$. In our case, we restrict the interval $[l_{min}, l_{max}]$ as the diagonal length of the input Cartesian radar image.
\bl{As demonstrated by \citet{ding2022translation}, utilizing sinograms with \ac{DFT} can provide both rotational and translational invariance. To achieve this, the \ac{RT} is used to generate sinograms for each radar image, and applying DFT to these sinogram images \eqref{eq:fft} ensures translational invariance.}
\begin{equation}
\centering
I_i(\theta,f) = \mathcal{F}(\mathcal{S}_i(\theta,l)) = \sum_{l} \mathcal{S}(\theta,l)\cdot e^{-2\pi i\left(\frac{fl}{M}\right)}
\label{eq:fft}
\end{equation}

\subsection{Radar Similarity Appraisal with Mutable Threshold}
The primary derivation of the distance score in our method is combining two techniques: cross-correlation and mutable threshold. 
Cross-correlation is commonly used to measure the similarity \bl{between two sets of data}. However, calculating the cross-correlation directly in image data can be time-consuming. Therefore, we utilize the frequency domain cross-correlation algorithm to generate the basis for the place similarity score.
The cross-correlation algorithm in frequency domain is described in \eqref{eq:fftcorr}.
\begin{equation}
\centering
C_{q,i} = \sum_{\theta}\mathcal{F}^{-1}(\mathcal{F}(I_q)\mathcal{F}^{*}(I_i))
\label{eq:fftcorr}
\end{equation}
We apply the \ac{DFT} to candidate and query matrices $I_i$ and $I_q$, and convert them into the frequency domain. The candidate and query matrices are constructed as lists of the discrete Fourier transformed vectors for each theta in sinogram $\mathcal{S}$ \eqref{eq:fft}. Next, the cross-power spectrum is computed by multiplying the \ac{DFT} of the query matrix with the complex conjugate of the \ac{DFT} of the candidate matrix. Finally, the inverse Fourier transform is applied to the cross-power spectrum data and summed along one axis, resulting in a one-dimensional array of cross-correlation $C_{q,i}$.

To account for differences between the query image and candidates, \bl{we calculate the auto-correlation value $C_{auto}$ as} a mutable threshold. Because the radar image captures the full surrounding environment, the score obtained from a single cross-correlation array may not always accurately reflect the information in the query radar image. To address this issue, we improve the standard of the similarity score by incorporating two cross-correlation results. Both the image similarity score for the auto-correlation and the query-candidate are determined by selecting the maximum value from the cross-correlation array. To identify the answer from the candidate images, \bl{we consider the candidate whose obtained score} is closest to the auto-correlation value.
Given that mutable query thresholds, the final place similarity distance $d$ is calculated according to \eqref{eq:distance}

\begin{equation}
\centering
d_i = |\max(C_{auto})-\max(C_{q,i})|
\label{eq:distance}
\end{equation}
\begin{equation}
\centering
X_q = \argmin_{i}(d_i)
\label{eq:target}
\end{equation}

The output of the place recognition model is represented by $X_q$ in \eqref{eq:target}, which is a candidate minimizing the similarity distance $d$.

\subsection{Place Information Retrieval}
Place decisions are conducted based on the place similarity distance, but as the interminable driving, the number of place candidates increases, leading to performance degradation. To enable efficient place retrieval across a wide range of fields without sacrificing performance, an efficient searching algorithm is required. To this end, we adopt a hierarchical method for agile minima access. Since the scores for neighboring scenes are analogous, we extract representative candidates from the cognate image collections. Under the assumption that resolution variances are irrelevant to the distance score tendency, we first obtain keyframe scores from low-resolution data, and then refine the candidate set using the highly rated group and neighboring data to achieve accurate place retrieval.


\section{experiment}
\label{sec:experiment}
\subsection{Radar Datasets}
To ensure the reliability of verification, we evaluate our algorithm on the two datasets: \textit{Oxford Radar Robotcar}~\cite{barnes2020oxford} and \textit{MulRan}~\cite{kim2020mulran}. \tabref{tab:dataset} illustrates a comparison of key characteristics for the dataset.
The \textit{Oxford} dataset is a short-term radar dataset that spans a week and covers the same area, making it suitable for straightforward intra-loop closure verification and consecutive global place recognition. In contrast, \textit{MulRan} is a long-term dataset that allows for temporal-global place recognition. \textit{MulRan} is pertinent for substantiating both single-session and multi-session loop closure, as each session drives different routes. We test our single-session loop closure algorithm on two \textit{Oxford} datasets and four locations in \textit{MulRan} and conduct multi-session verification in each representative location.
\begin{table}[h!]
\resizebox{\linewidth}{!}{
\begin{tabular}{cc|cccc}
\multicolumn{2}{c}{\textbf{Dataset}}                    & \textbf{Distance} & \textbf{Max. Interval} & \textbf{Sequences} & \textbf{Sessions} \\ \hline\hline
\multicolumn{1}{c}{\multirow{4}{*}{MulRan}} & DCC       & 4.9km             & 33 days                        & 3                  & Intra             \\ 
\multicolumn{1}{c}{}                        & KAIST     & 6.1km             & 75 days                        & 3                  & Intra             \\ 
\multicolumn{1}{c}{}                        & Riverside & 6.8km             & 22 days                        & 3                  & Intra/Multi       \\ 
\multicolumn{1}{c}{}                        & Sejong    & 23.4km            & 62 days                        & 3                  & Multi             \\ \hline
\multicolumn{2}{c|}{Oxford}               & 9km               & 8 days                         & 32                 & Intra/Multi       \\ \hline
\end{tabular}}
\caption{Radar Dataset Attributes}\label{tab:dataset}
\vspace{-3mm}
\end{table}

\begin{figure}[!t]
    \centering
    \includegraphics[width=\columnwidth]{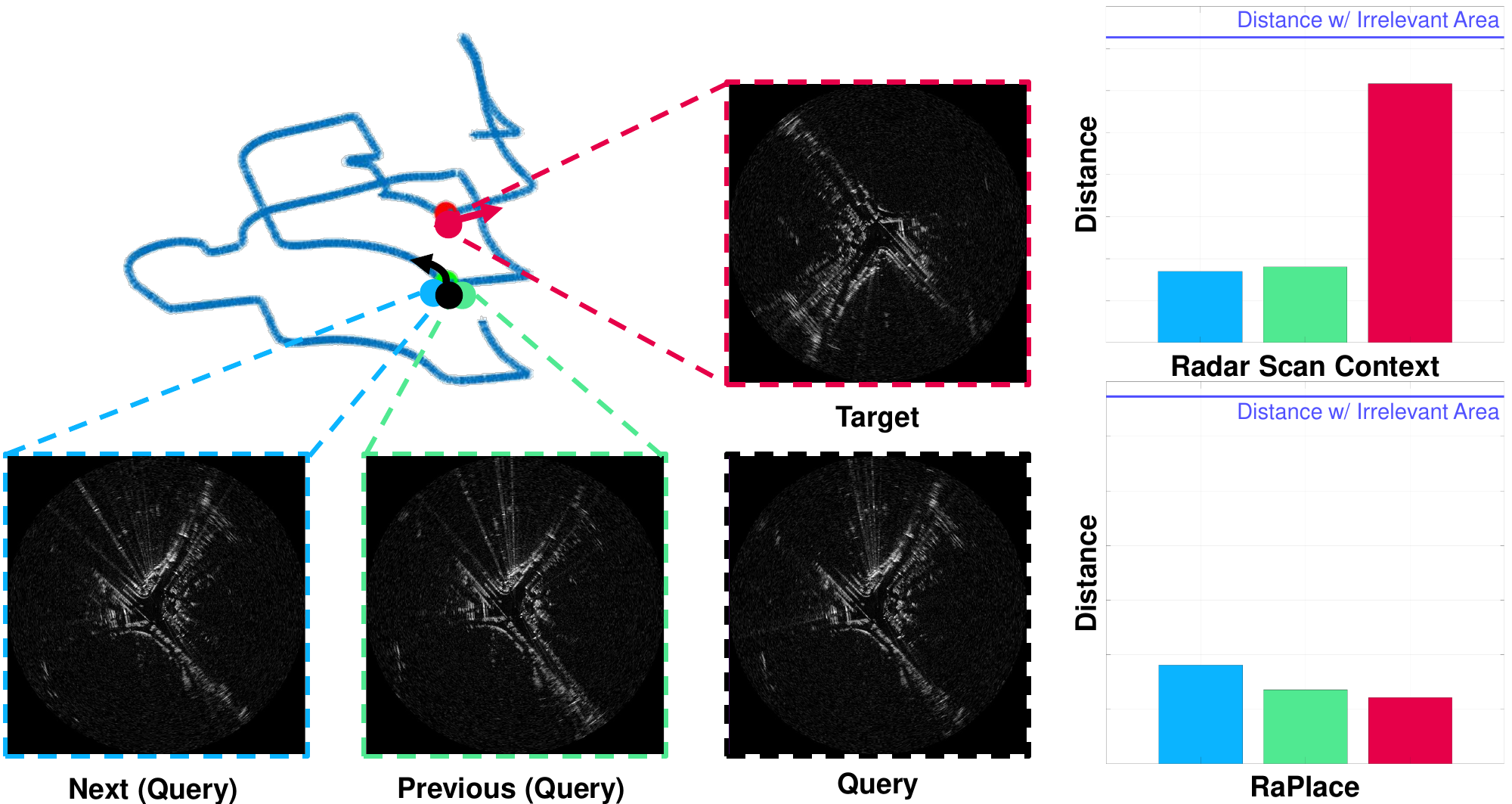}
    \caption{Time elevation graph and place matching example in the KAIST sequence. While both Radar Scan Context and RaPlace typically indicate high similarity between neighboring frames, there are certain scenarios where they exhibit different behavior. For instance, when the vehicle undergoes a full rotation and there is a distance of approximately 8 meters between the two positions, Radar Scan Context detects a substantial difference from the neighboring frames, whereas RaPlace remains stable.}
    \label{fig:kaist_ex}
    \vspace{-5mm}
\end{figure}
\subsection{Evaluation Criteria}

\subsubsection{True Positives Matching Graph}
Demonstrating the indisputable accuracy of our place recognition approach, we visualize the matching results for each position. Results indicate the true-positive predictions for each location and demonstrate the robustness and stability of the algorithm. Furthermore, the graphs reveal the vulnerability of the place recognition descriptor. 
By creating a true-positive point trajectory for the four intra-sessions and TP detection occurrence graphs for the three multi-sessions, we compare all matching results with the state-of-the-art method for structural radar place recognition \textit{MulRan} (Radar Scan Context)~\cite{kim2020mulran}. We evaluate the performance of our method and assess the effectiveness in place recognition.

\subsubsection{Assessment with Scores}
In our evaluation, we configure the place identification standards to operate within a 20-meter range, accounting for the sparsity of the imaging radar data. An alleviated matching standard is recommended for radar place recognition in contrast to LiDAR, which is better suited for detecting rough environmental changes.
To evaluate the performance of our prediction model, we rely on the Area Under the Curve (AUC) value from the \ac{PR}-curve as a reliable criterion for evaluation. Additionally, we account for the F1 score-recall curve and the maximum F1 score for a comprehensive assessment of our method. When evaluating the prediction model, we calculate Recall@1 only when the exact answer exists among the candidates. However, since the matching answer may not always exist in SLAM, Recall@1 is not a practical measure for scarce loop closure instances.

\begin{table}[t!]
\centering
\resizebox{0.8\columnwidth}{!}{
\begin{tabular}{cc|c|cc}
\hline
\multicolumn{2}{c|}{\multirow{2}{*}{\textbf{Dataset}}}                    & \multirow{2}{*}{\textbf{Method}} & \multicolumn{2}{c}{\textbf{Single Session}}\\
\multicolumn{2}{c|}{}                                                     &                                  & \multicolumn{1}{c}{\textbf{AUC}}   & \textbf{F1 Score}\\ \hline \hline
\multicolumn{1}{c|}{\multirow{8}{*}{MulRan}} & \multirow{2}{*}{DCC}       & R-SC                               & \multicolumn{1}{c}{0.801}          & 0.679            \\
\multicolumn{1}{c|}{}                        &                            & RP                               & \multicolumn{1}{c}{\textbf{0.843}} & \textbf{0.735}   \\ \cline{2-5}
\multicolumn{1}{c|}{}                        & \multirow{2}{*}{KAIST}     & R-SC                               & \multicolumn{1}{c}{0.819}          & \textbf{0.786}   \\
\multicolumn{1}{c|}{}                        &                            & RP                               & \multicolumn{1}{c}{\textbf{0.881}} & 0.781            \\ \cline{2-5}
\multicolumn{1}{c|}{}                        & \multirow{2}{*}{Riverside} & R-SC                               & \multicolumn{1}{c}{0.944}          & 0.862            \\
\multicolumn{1}{c|}{}                        &                            & RP                               & \multicolumn{1}{c}{\textbf{0.959}} & \textbf{0.897}   \\ \cline{2-5}
\multicolumn{1}{c|}{}                        & \multirow{2}{*}{Sejong}    & R-SC                               & \multicolumn{1}{c}{0.904}          & 0.830            \\
\multicolumn{1}{c|}{}                        &                            & RP                               & \multicolumn{1}{c}{\textbf{0.924}} & \textbf{0.858}   \\ \hline
\multicolumn{1}{c|}{\multirow{4}{*}{Oxford}} & \multirow{2}{*}{19-01-16}  & R-SC                               & \multicolumn{1}{c}{0.545}          & 0.534            \\
\multicolumn{1}{c|}{}                        &                            & RP                               & \multicolumn{1}{c}{\textbf{0.595}} & \textbf{0.543}   \\ \cline{2-5}
\multicolumn{1}{c|}{}                        & \multirow{2}{*}{19-01-18}  & R-SC                               & \multicolumn{1}{c}{0.826}          & \textbf{0.773}   \\
\multicolumn{1}{c|}{}                        &                            & RP                               & \multicolumn{1}{c}{\textbf{0.827}} & 0.734            \\ \hline
\end{tabular}
}
\caption{Intra-session Place Recognition Results}\label{tab:pr_result}
\vspace{-5mm}
\end{table}

\begin{figure*}[!t]
\centering
    \begin{minipage}{0.67\textwidth}
    \centering
    \subfloat[MulRan DCC\label{fig:pr_dcc}]{
		\includegraphics[bb = 20 200 600 600 ,clip=true, width=0.32\textwidth]{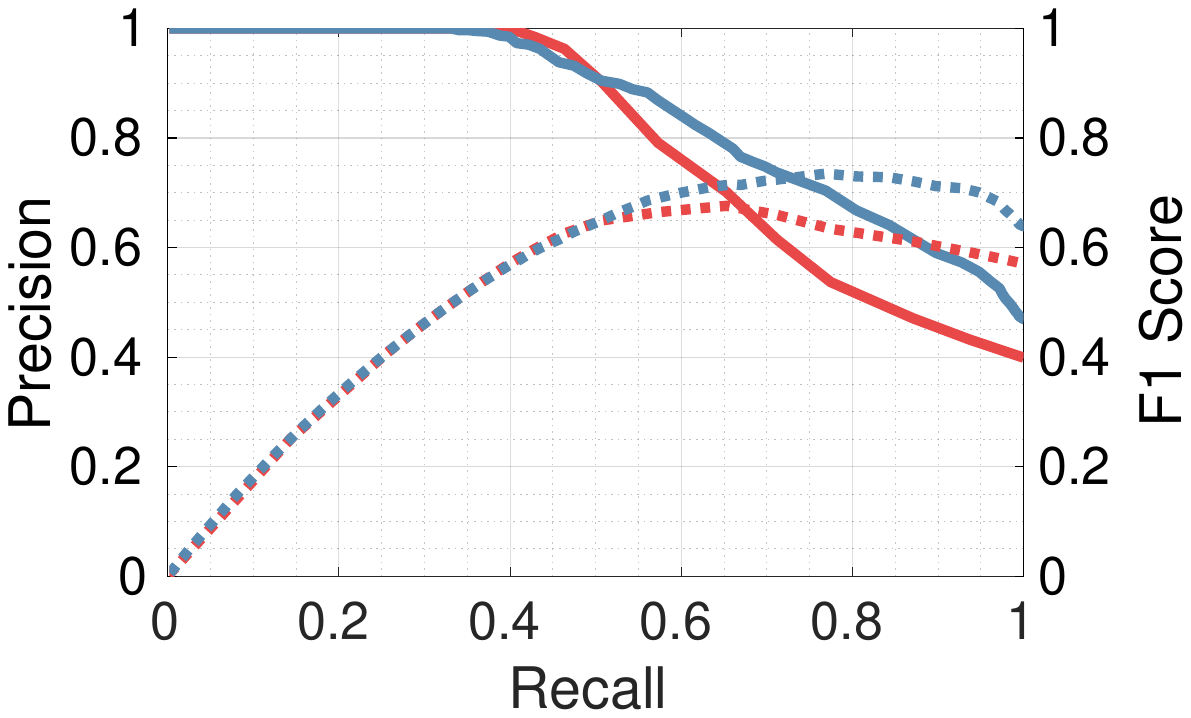}
	}
    \subfloat[MulRan DCC\label{fig:ele_DCC}]{
		\includegraphics[width=0.32\textwidth]{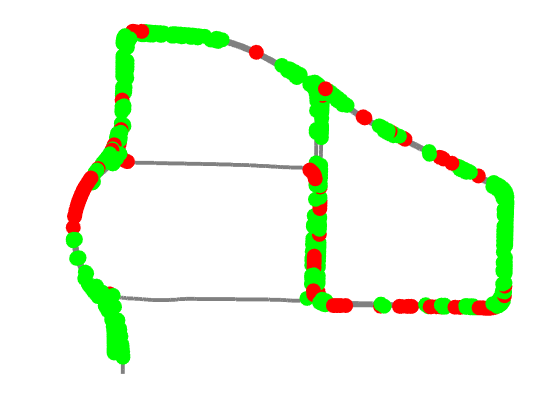}
	}
    \subfloat[MulRan KAIST\label{fig:ele_KAIST}]{
		\includegraphics[width=0.32\textwidth]{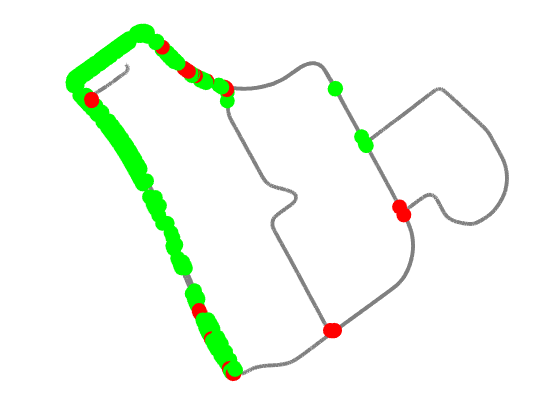}
	}\hfil
 \subfloat[Oxford 2019-01-16\label{fig:pr_ox}]{
		\includegraphics[bb = 20 200 600 600 ,clip=true, width=0.32\textwidth]{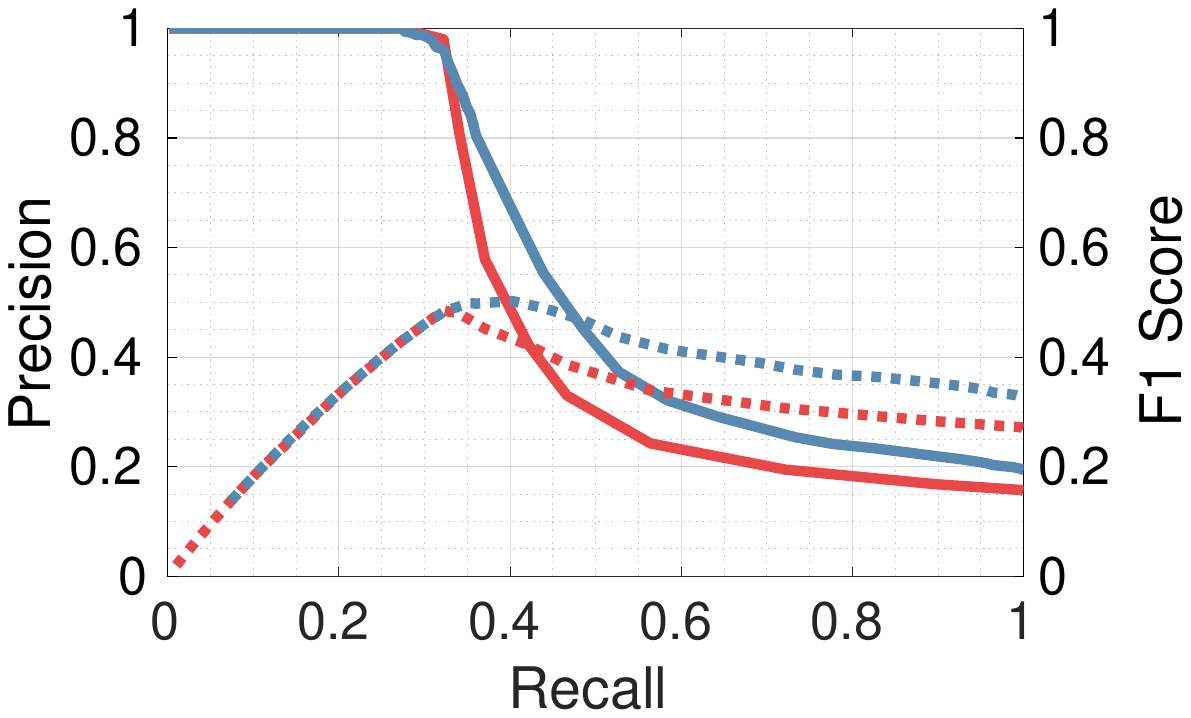}
	}
	  \subfloat[Oxford 2019-01-16\label{fig:ele_ox}]{
		\includegraphics[width=0.32\textwidth]{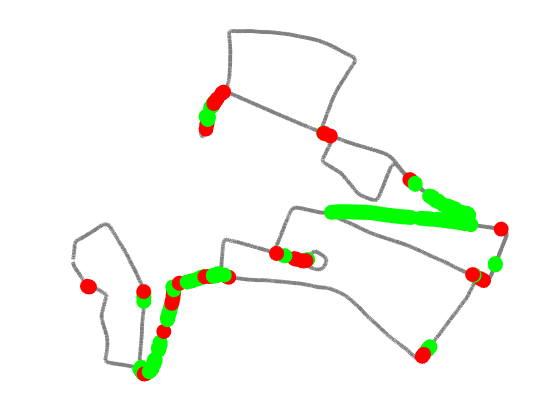}
	}
    \subfloat[MulRan Riverside\label{fig:ele_river}]{
		\includegraphics[width=0.32\textwidth]{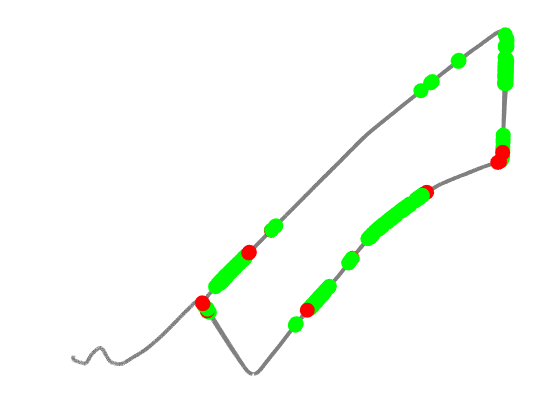}
	}
\vspace{3mm}
    \end{minipage}
    \begin{minipage}{0.24\textwidth}
    \centering
    \subfloat[\label{fig:pr_bar}]{
		\includegraphics[bb = 80 40 500 780 ,clip=true,width=\textwidth]{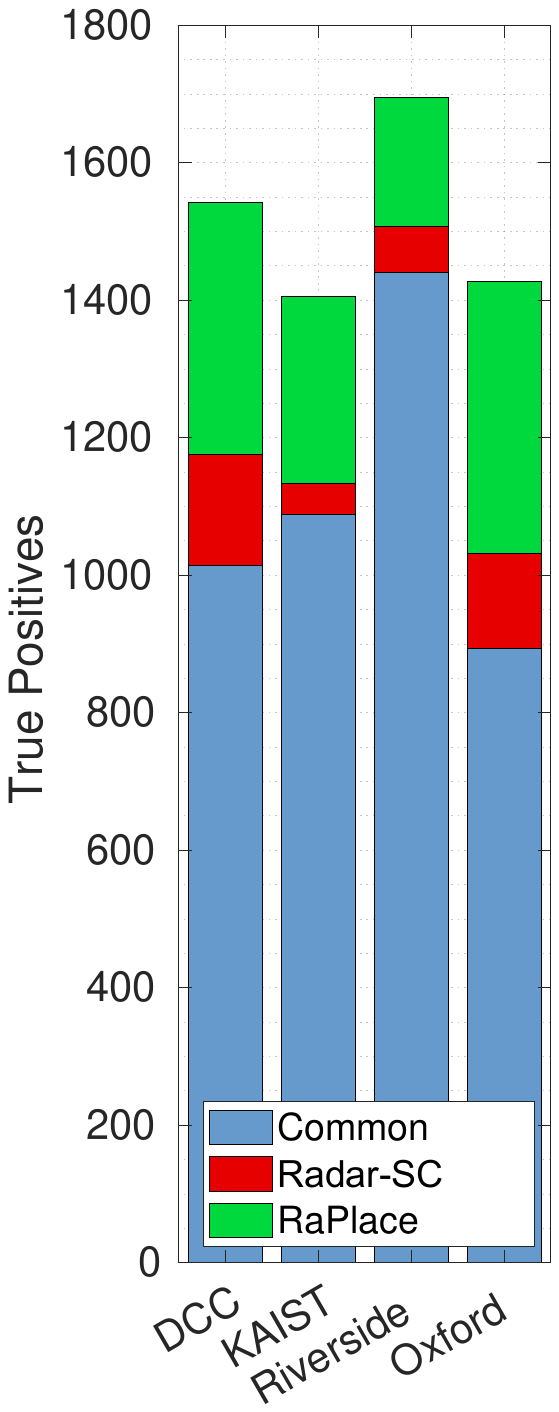}
	}
\end{minipage}
    \vspace{-4mm}
  	\caption{The figure on the left illustrates the \ac{PR} and F1-recall curves for the intra-session evaluation. The solid lines represent the precision values, while the dotted lines represent the F1 score for each recall. The results obtained by the proposed method (blue line) are substantially superior to those of Radar Scan Context (red line), with the proposed method detecting more latent links that improve the precision values at high recalls. However, for the Oxford dataset, the low precision scores for both methods are attributed to the scarcity of revisited locations. The middle part of the figure depicts the trajectory, with points where only the proposed method (green) or Radar Scan Context (red) detect the same place. The statistical graph on the right indicates that our proposed method detects more true positive results than the Radar Scan Context.}
	\label{fig:pr_result}	
\end{figure*}

\begin{figure*}[!t]
 \centering
    \subfloat[]{
	 	\includegraphics[width=0.7\textwidth]{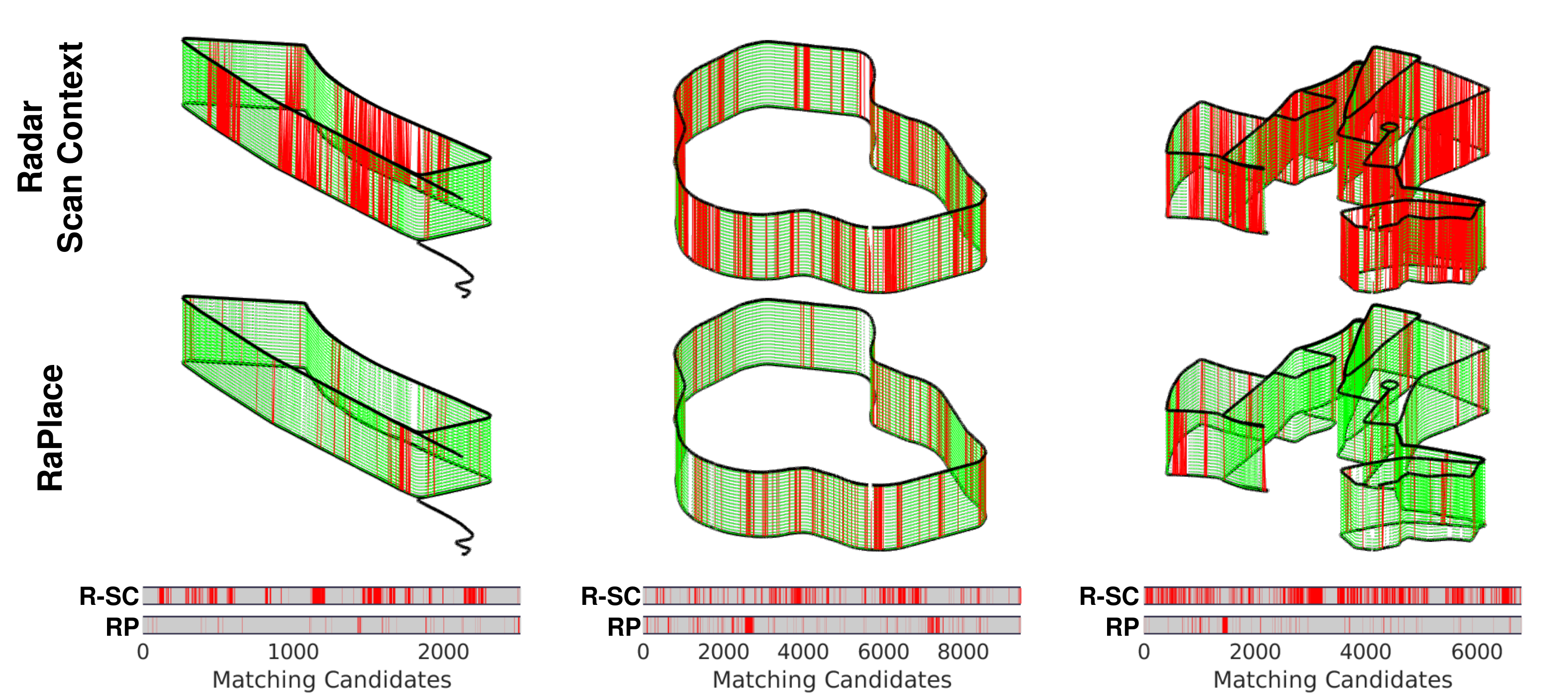}
	   }
    \subfloat[]{
    \begin{minipage}[b]{0.29\textwidth}
        \centering
        \hspace{8pt}
        \resizebox{0.8\textwidth}{!}{
        \centering
        \begin{tabular}{c|cc} \toprule
            \multicolumn{1}{c|}{\multirow{2}{*}{\textbf{Dataset}}} &  \multicolumn{2}{c}{\textbf{TP-Detection Rate}} \\
           & Radar-SC & RaPlace\\ \midrule
          Riverside & 0.671& \textbf{0.938}   \\
          Sejong &0.676 & \textbf{0.812}   \\
          Oxford & 0.485  & \textbf{0.814}   \\
          \bottomrule
          \end{tabular}%
          }
      \hfill
      \vspace{10pt}
        \resizebox{\textwidth}{!}{
        \begin{tabular}{cc|c|cc}
        \hline
        \multicolumn{2}{c|}{\multirow{2}{*}{\textbf{Dataset}}}                    & \multirow{2}{*}{\textbf{Method}} & \multicolumn{2}{c}{\textbf{Global PR}}       \\
        \multicolumn{2}{c|}{}                                                     &                                  & \multicolumn{1}{c}{\textbf{AUC}}   & \textbf{F1 Score} \\ \hline \hline
        \multicolumn{1}{c|}{\multirow{8}{*}{MulRan}} & \multirow{2}{*}{DCC}       & R-SC                               & \multicolumn{1}{c}{0.952}          & 0.867             \\
        \multicolumn{1}{c|}{}                        &                            & RP                               & \multicolumn{1}{c}{\textbf{0.978}} & \textbf{0.932}    \\ \cline{2-5}
        \multicolumn{1}{c|}{}                        & \multirow{2}{*}{KAIST}     & R-SC                               & \multicolumn{1}{c}{0.978}          & 0.899             \\
        \multicolumn{1}{c|}{}                        &                            & RP                               & \multicolumn{1}{c}{\textbf{0.984}} & \textbf{0.930}    \\ \cline{2-5}
        \multicolumn{1}{c|}{}                        & \multirow{2}{*}{Riverside} & R-SC                               & \multicolumn{1}{c}{0.915}          & 0.802             \\
        \multicolumn{1}{c|}{}                        &                            & RP                               & \multicolumn{1}{c}{\textbf{0.995}} & \textbf{0.971}    \\ \cline{2-5}
        \multicolumn{1}{c|}{}                        & \multirow{2}{*}{Sejong}    & R-SC                               & \multicolumn{1}{c}{0.929}          & 0.819             \\
        \multicolumn{1}{c|}{}                        &                            & RP                               & \multicolumn{1}{c}{\textbf{0.969}} & \textbf{0.906}    \\ \hline
        \multicolumn{1}{c|}{\multirow{2}{*}{Oxford}} & \multirow{2}{*}{19-01-16}  & R-SC                               & \multicolumn{1}{c}{0.773}          & 0.679             \\
        \multicolumn{1}{c|}{}                        &                            & RP                               & \multicolumn{1}{c}{\textbf{0.959}} & \textbf{0.949}    \\ \hline
        \end{tabular}
        }
        \vspace{1pt}
    \end{minipage}
    }
    \caption{The global place recognition results for three large-scale sequences. We represent all matching results obtained by the Radar Scan Context and RaPlace as green lines and the detection failures of each method as red lines. The Radar Scan Context fail to detect certain places due to the rigid transformation information being different from the original path. However, our proposed method is robust in detecting the surrounding places, resulting in a high rate of true positive detection.}
    \label{fig:pr_gloc_result}	
    \vspace{-5mm}
\end{figure*}

\subsection{Intra-session Link Proposal}
The DCC, KAIST, and Riverside datasets from \textit{MulRan} and each sequence from the \textit{Oxford} dataset are rich in loop closures, making them suitable for verifying the reverse lane recognition and data attributes in our method.
The intra-session place recognition evaluation results or Radar Scan
Context (R-SC) and RaPlace (RP) are presented in \figref{fig:intro}, \figref{fig:pr_result} and \tabref{tab:pr_result}, where the true-positive match consequences are identified, allowing us to detect where the localization performances are refined. 

To visually represent the results of our evaluation, we create a trajectory graph that features true-positive points. This graph includes green points that denote detections that our algorithm identifies but are missed by Radar Scan Context and red points that represent detections that our algorithm fails to detect but are identified by Radar Scan Context. This approach allows us to easily compare the performance of our method to the state-of-the-art method for structural radar place recognition.
As shown in \figref{fig:pr_result}, the proposed method finds a significant number of true-positive matches that the Radar Scan Context overlooked. Notably, our algorithm is particularly robust in places where both rotational and translational intervals exist. 

The quantitative results in \figref{fig:pr_result} and \tabref{tab:pr_result} include specific numerical values that indicate the precision, F1 score, and recall of the prediction models. Compared to existing methods, the proposed method shows a relaxed decrease in precision. RaPlace exhibits invariance for image information in matching, which allows it to identify obscure place pairs and increase overall precision. The proposed method outperforms existing structural methods in terms of the \ac{AUC}. Since the precision is supplemented for the ample recall in our algorithm, the maximum value of the F1 score is extended and located at a higher recall value.

\subsection{Multi-session Place Recognition}
\label{sec:multi_result}
The results of the multi-session evaluation are presented in \figref{fig:pr_gloc_result}, and they are consistent with the intra-session results. In the multi-sequence trajactory graphs, green lines represent matched links between two sequences using all place recognition methods, while red lines denote instances where each method failed to detect a match.
Our proposed method outperforms Radar Scan Context regarding detection rates, with fewer missed detections overall. As shown in the table in \figref{fig:pr_gloc_result}, the density of entire true-positive detections is significantly improved with the proposed method, demonstrating accuracy in global place recognition.
The quantitative global place recognition results for all sequences are summarized in the tables in \figref{fig:pr_gloc_result}. We achieve comprehensive precision values in adequate recalls by reducing the false-positive ratio. The performance of global place recognition is significantly improved by increasing the absolute true-positive quantity. 


\begin{figure}[!t]
   \centering
   \vspace{-3mm}
   \hspace{-3mm}
   \subfloat[Intra-Session\label{fig:ab_sc_rot}]{
		\includegraphics[bb = 20 200 540 570 ,clip=true,width=0.48\columnwidth]{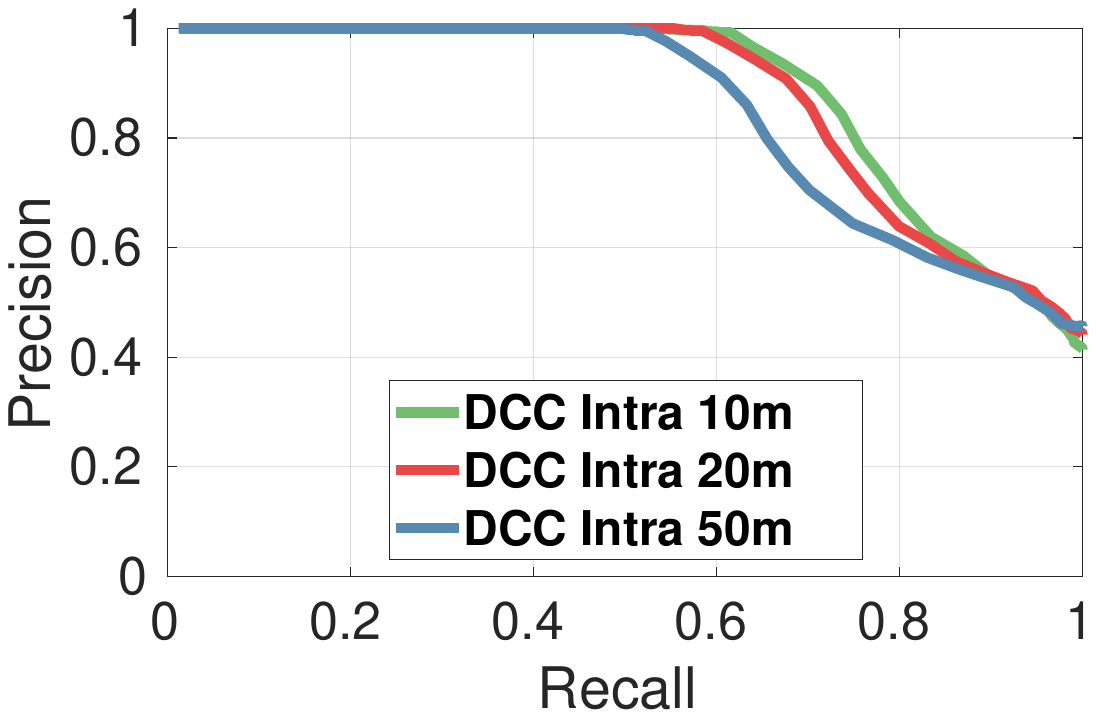}
	}
	\subfloat[Global Place Recognition\label{fig:ab_rp_rot}]{
		\includegraphics[bb = 20 200 540 570 ,clip=true,width=0.48\columnwidth]{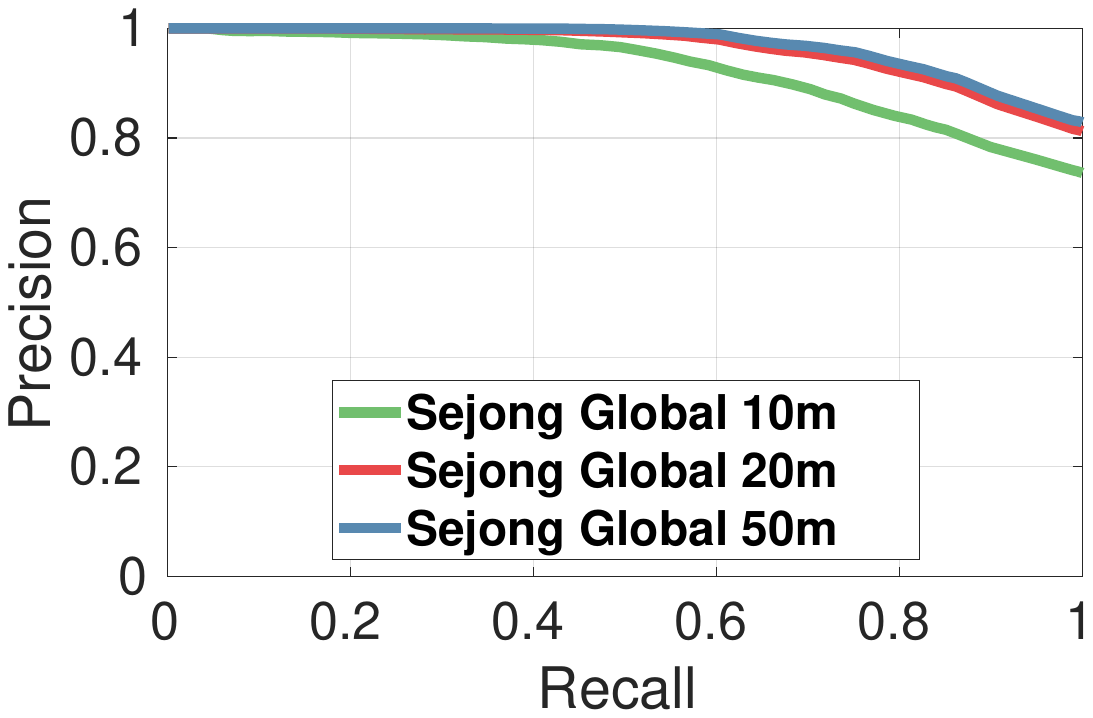}
	}
   \caption{The figure illustrates the results for the variance of the loop detection criteria, which are set at 10m, 20m, and 50m. For the intra-session evaluation, the precision values decrease as the criteria was relaxed. However, for the global place recognition results, expanding the criteria improves the detection accuracy.}
   \label{fig:abla}
   \vspace{-5mm}
\end{figure}

\subsection{Effect of Loop Detection Criteria}
When conducting multi-session place recognition, we observe that the rearranged driving route and the temporal difference lead to confusion in searching for the same location. Therefore, ground truth boundary configuration is essential to evaluate the performance of the global place recognition algorithm, which must take into account the different traffic lanes containing opposite positions. We obtain the prediction results for the 10, 20, and 50 meters boundary and depict \ac{PR}-curves for the DCC and Sejong from \textit{MulRan}.
\figref{fig:abla} is the \ac{PR} curves for the multiple ground truth boundaries. Our evaluation indicates that in the intra-session case, a larger ground truth boundary leads to less accurate results. This is because intra-session sequences share nearly identical environments, readily finding accurate correspondences. Additionally, intra-session sequences typically follow short paths, which limits the number of candidates that need to be considered.
However, we observe that alleviating the ground truth boundary improves the accuracy of the global place recognition. In global place recognition, we need to search a large amount of data to find suitable candidates, and the environments in the candidate data may differ from the query data. In this process, an alleviated ground truth boundary helps to identify nearby environments and find the most similar representations from the candidate data.
These findings support our initial assumptions regarding the significance of radar frequency and lane width for place recognition criteria. Also, the radar range is up to 200m, mitigated confines of ground truth represent desirable candidates for loop closure, resulting in reliable true-positive results.

\subsection{Sensitivity of Place Recognition}
To assess the robustness of our proposed model, we evaluate our place recognition algorithm to verify sensitivity. Although we claim to achieve both rotational and translational invariance, it is imperative to provide empirical evidence to support our assertion. To compare the performance of Radar Scan Context and RaPlace, we conduct a graph tendency analysis subject to different thresholding standards. We select a query sample from the radar dataset and generate candidate samples by rotating and rearranging both vertical and horizontal directions. To avoid squandering neighbor radar information, we crop the query image from the original full Cartesian image. Each rotated and translated image is occupied with real surrounding information and devoid of empty black spaces. 

\begin{figure}[!t]
   \centering
   \vspace{-3mm}
   \includegraphics[width=\columnwidth]{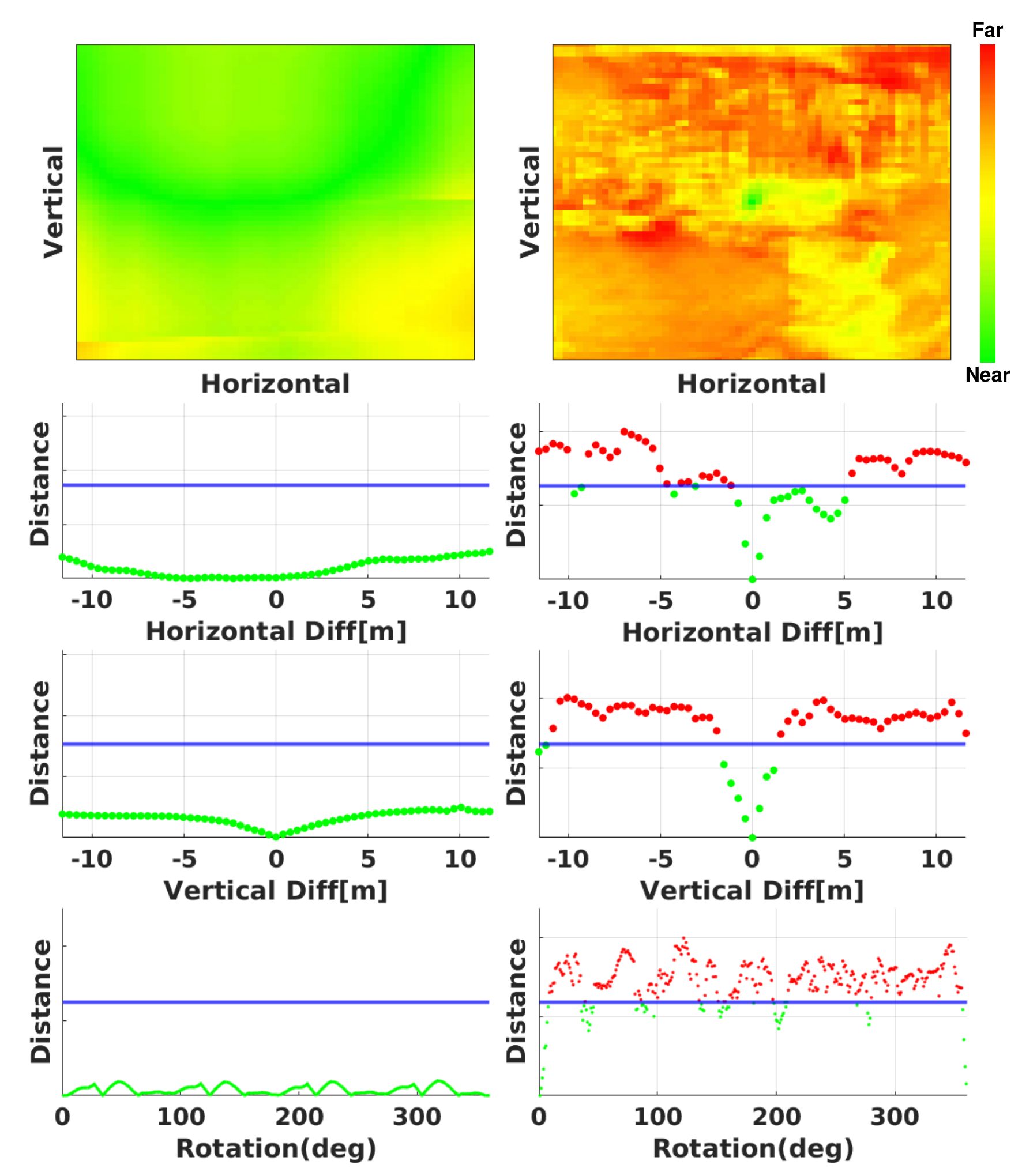}
   \caption{The sensitivity evaluation results for virtual candidates, which are generated by rotating and cropping the query image. The distance scores are measured to verify the algorithm's coverage. Across all variations, RaPlace exhibits a more consistent and stable distance score than Radar Scan Context. While all distance values have to be below the threshold (Blue line), Radar Scan Context only provides a reliable score for a limited area.}
   \label{fig:ablation}
   \vspace{-5mm}
\end{figure}

\figref{fig:ablation} displays the results of the similarity distance analysis. Radar Scan Context and RaPlace are used to determine the minimum distance to assure place analogy, and the algorithms are relative detection methods. To ensure a fair comparison of the performance of each algorithm, we normalize all graphs by reflecting the threshold values of each algorithm. The threshold values are calculated using the two radar data from completely different places. We observe that Radar Scan Context has a reliable score in the periphery of minima for both rotational and translational variance. However, it requires precise revisits to achieve confident place recognition, and analogous places could not attain a definitely distinguishable score. Conversely, the proposed method exhibits robustness for both rotational and translational variance, with all our scores remaining below the threshold value. Our sensitivity study provides compelling evidence as to why the RT-based algorithm identifies more places and offers inclusive localization candidates.

\subsection{Computational Time}
We evaluate the execution time of the proposed algorithm, which consists of three primary steps: polar to Cartesian conversion, Radon transform, and image retrieval. \bl{The statistical computations were performed utilizing a 2.5GHz 16-Core Intel i7-11700 CPU. Retrieving 2000 frames took approximately 30ms, achieving real-time performance.} The data loading time for all DCC data is approximately 38.4ms, and given that radar images are produced at a rate of 4Hz, our algorithm can process the offline dataset in real-time.  However, the average computing time for the polar to Cartesian conversion and the Radon transform are about 85.25ms and 86.58ms, respectively, making it challenging to ensure real-time performance in long-term online environments.
\section{Conclusion}
\label{sec:conclusion}
We propose a novel radar place recognition algorithm that is rigid transformation-invariant and noise-irrelevant. Our methodology involves the computation of similarity scores and the estimation of similarity distances through \bl{the use of a mutable threshold}. Results derived from experimentation and ablation studies indicate that our method produces promising outcomes.
Despite our methodology utilizing \ac{RT} and \ac{DFT} to condense image information, the computational time requires to process high-resolution images still presents a challenge. As such, the development of a time-image resolution complementary algorithm is imperative to enable efficient storage and retrieval of radar location information. 
As part of our future work, we plan to modify our algorithm to be compatible with radar-based modules. Given that our method produces a greater number of potential candidates for loop closure than preexisting radar place recognition methodologies, we expect that our approach could be leveraged to achieve highly accurate radar-based \ac{SLAM}.

\balance
\small
\bibliographystyle{IEEEtranN} 
\bibliography{string-short,references}

\end{document}